\documentclass{article}

\usepackage[nonatbib, final]{neurips_data_2024}

\usepackage[utf8]{inputenc} 
\usepackage[T1]{fontenc}    
\usepackage{hyperref}       
\usepackage{url}            
\usepackage{booktabs}       
\usepackage{amsfonts}       
\usepackage{nicefrac}       
\usepackage{microtype}      
\usepackage{xcolor}         
\usepackage{graphicx}
\usepackage{subcaption}
\usepackage{multirow}
\usepackage{array}
\usepackage{amsmath}
\usepackage{amssymb}
\usepackage{amsthm}
\usepackage[toc,page]{appendix}
\title{Generated Indoor video frames for Texture-less point tracking}

\author{
  Jianzheng Huang\textsuperscript{\rm 1}\thanks{Equal contribution.} \quad Xianyu Mo\textsuperscript{\rm 1}\footnotemark[1] \\ 
\textbf{Ziling Liu}\textsuperscript{\rm 1,2}\quad \textbf{Jinyu Yang}\textsuperscript{\rm 2} \quad \textbf{Feng Zheng}\textsuperscript{\rm 1}\thanks{Corresponding author.}
  \\
  \textsuperscript{\rm 1} Southern University of Science and Technology \quad
  \textsuperscript{\rm 2} tapall.ai\\
}

\begin{document}
\newcommand{\ziling}[1]{\textcolor{green}{#1}}
\maketitle

\begin{abstract}
Point tracking is becoming a powerful solver for motion estimation and video editing.
Compared to classical feature matching, point tracking methods have the key advantage of robustly tracking points under complex camera motion trajectories and over extended periods. However, despite certain improvements in methodologies, current point tracking methods still struggle to track any position in video frames, especially in areas that are texture-less or weakly textured. In this work, we first introduce metrics for evaluating the texture intensity of a 3D object. Using these metrics, we classify the 3D models in ShapeNet \cite{shapenet} into three levels of texture intensity and create GIFT, a challenging synthetic benchmark comprising 1800 indoor video sequences with rich annotations. Unlike existing point-tracking datasets that set ground truth points arbitrarily within scenes, all ground truth points in GIFT are specifically set to the classified target objects, ensuring that each video corresponds to a particular texture intensity level. Furthermore, we comprehensively evaluate current methods on GIFT to assess their performance across different texture intensity levels and analyze the impact of texture on point tracking. 
\end{abstract}

\begin{figure}[t]
    \centering
    \includegraphics[width=\textwidth]{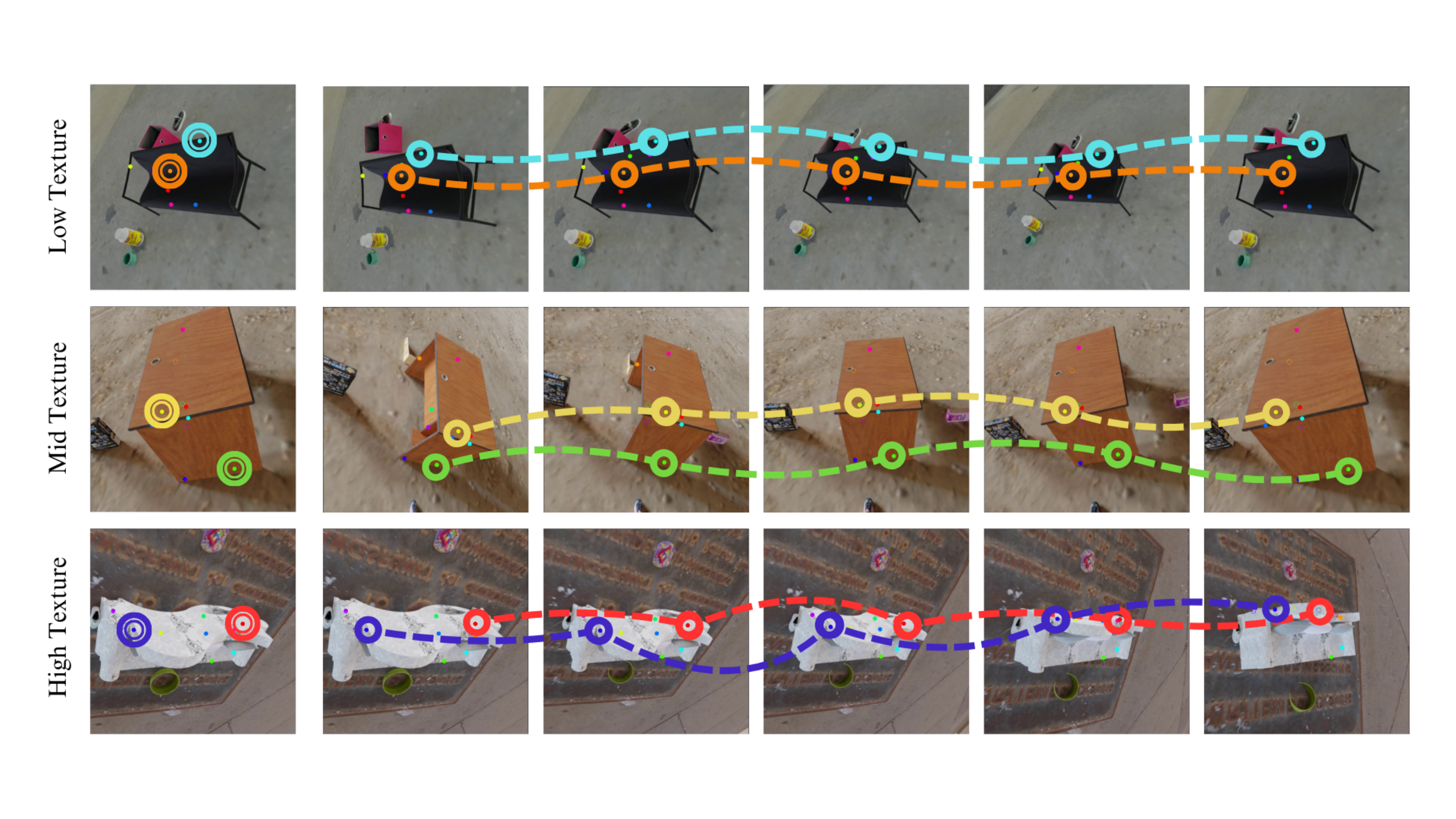}
    \caption{Visualization of video frames showing target objects of the same category with varying texture intensities. The video sequences from top to bottom illustrate weak, medium, and strong texture intensities, respectively.
}
    \label{fig:1}
\end{figure}
\section{Introduction}
Point tracking is a fundamental task in computer vision with numerous downstream applications, such as video object segmentation\cite{sam-pt, surgical-sam}, video editing\cite{videoswap, videoshop, videoMotion, revideo, mft, advert3D} and Motion estimation\cite{motionDrag, motionI2V, motionIpose}. Given a video and designated query points within a specific frame, the objective of point tracking is to continuously track these query points across the subsequent frames. 
Although current methods\cite{TAPVid, pips, tapir, omnimotion, mft, BootsTAP, spatialtracker} have made significant improvements in handling long videos and occlusions, their performance deteriorates when tracking points in texture-less regions. Existing datasets\cite{TAPVid, pointodyssey} arbitrarily set ground truth points within scenes without explicitly defining and differentiating textures. In a video sequence, ground truth points are distributed across objects with varying texture intensities. This ambiguous distribution makes it challenging to study the impact of texture on point tracking performance, especially in texture-less regions, due to the predominance of high-intensity textures.

Texture in an image refers to the visual patterns that represent the surface characteristics of an object. It can be described using terms such as smooth, rough, silky, or bumpy. Texture can be observed at both macroscopic and microscopic scales, providing crucial information about the structural arrangement of surfaces and their relationship to the surrounding environment. In image processing and analysis, texture is often vital for differentiating between objects with similar colors or shapes.
To comprehensively study how texture intensity impacts point-tracking methods, we make the first attempt to design metrics for evaluating 3D object texture. Specifically, we begin by rendering images of the 3D model from various perspectives. From these images, we extract Local Binary Pattern (LBP) \cite{LBP} features, count the number of FAST \cite{fast} keypoints, measure the RGB variance, and calculate the low-high frequency to comprehensively evaluate the texture intensity of the 3D model's surface. Using these metrics, we classify objects in ShapeNet\cite{shapenet} into three different relative texture intensity levels, as illustrated in Figure \ref{fig:1}. With these classified objects and indoor scenes collected from the internet, we utilize Kubric\cite{kubric} to create an indoor video dataset GIFT containing 1800 video sequences, where ground truth points are artificially set on the target objects. Notably, the target objects within each video share the same texture intensity level, ensuring that each video corresponds to a specific texture intensity. GIFT consists of 6 subsets, each representing different levels of tracking difficulty. These subsets are created based on the combination of two camera trajectory complexities and three texture intensities. The statistics of GIFT and comparison with other datasets are presented in Table~\ref{tab:dataset}.

We evaluated a series of recent point tracking methods and a state-of-the-art optical flow method on GIFT. Based on the experimental results, we made some intriguing observations. 1) The accuracy of point tracking significantly diminishes as the texture intensity in the tracking area decreases. This underscores the inherent challenge of accurately locating and tracking specific pixels in regions with weak textures. 2) When the texture intensity remains constant and the camera trajectory becomes more complex, point-tracking methods with a global receptive field can potentially enhance tracking performance slightly. This offers a promising direction for addressing the challenge of weak texture point tracking.

Overall, our contributions are threefold:

\begin{itemize}
    \item We introduce novel metrics to evaluate the texture intensity of 3D models which provides a detailed assessment of texture variations, crucial for studying point tracking performance in different texture regions.
    \item We developed a camera trajectory generation method and utilized Kubric\cite{kubric} to synthesize a challenging dataset with every video corresponding to a specific camera trajectory complexity and texture intensity.
    \item We evaluated several recent point tracking methods on GIFT and discovered two interesting findings that offer valuable insights for future improvements in point-tracking algorithms.
\end{itemize}

\begin{table}[t]
\centering
\caption{GIFT statistics and comparison with other datasets.}
\label{tab:dataset}
\resizebox{\textwidth}{!}{%
\begin{tabular}{>{\raggedright\arraybackslash}p{3.5cm}|ccccccc}
\toprule
& MPI Sintel\cite{MPI} & Flyingthings++\cite{flying}& Kubric\cite{kubric} & TAP-Vid-Kinetics\cite{kinetics} & TAP-Vid-DAVIS\cite{davis2017} &PointOdyssey\cite{pointodyssey}  &GIFT\\ \midrule
Resolution & 436 $\times$ 1024 & 540 $\times$ 960 & 256 $\times$ 256 & $\geq$ 720 $\times$ 1280 & 1080 $\times$ 1920 &540 $\times$ 960  &512  $\times$512\\ 
Avg. trajectory count & 436 $\times$ 1024 & 1,024 & Flexible & 26.3 & 21.7 &18,700  &Flexible\\ 
Avg. span of trajectories & 4\% & 100\% & 100\% & 30\% & 30\% &100\%  &100\%\\ 
Avg. frames per video & 50 & 8 & 24 & 250 & 67 &2,035  &60\\ 
Training frames & 1064 & 21818 & Flexible & - & - &166K  & - \\ 
Validation frames & - & 4248 & - & - & - &24K  & -\\ 
Test frames & 564 & 2247 & - & 297K & 1999 &26K  & 108K\\ 
Texture annotations &$\times$ &$\times$ &$\times$ &$\times$ &$\times$ &$\times$ &$\checkmark$\\
Controllable annotation regions & $\times$ & $\times$ & $\times$ & $\times$ & $\times$ &$\times$  &$\checkmark$\\
Scene randomization & $\times$ & $\checkmark$ & $\checkmark$ & $\times$ & $\times$ &$\checkmark$  &$\checkmark$\\ 
Multiple views & $\times$ & $\times$ & $\checkmark$ & $\times$ & $\times$ &$\checkmark$  &$\checkmark$\\ 
Continuous & $\times$ & $\times$ & $\checkmark$ & $\checkmark$ & $\checkmark$ &$\checkmark$  &$\checkmark$\\ 

\bottomrule
\end{tabular}
}
\end{table}


\section{Related Work}
\subsection{Point Tracking Methods}
The methods for constructing point correlations across video frames can be mainly divided into two categories: optical flow and point tracking. Optical flow methods \cite{flownet, raft, loftr, videoflow} aims at estimating the motion of every pixel between video frames. These models track each point by accumulating pixel displacement for each frame which makes it very prone to tracking drift and unable to handle occlusion situations. Point tracking methods\cite{TAPVid,pips,mft,pointodyssey,cotracker,track_everything} aim at tracking sparse target points more dedicatedly and robustly. Inspired by cost volume \cite{costVolume}, Doersch \textit{et al}.\cite{TAPVid} proposed the first end-to-end deep learning algorithm to track any point in a video. Harley \textit{et al} \cite{pips}. introduced a method that employs 4D cost volume \cite{4DCostVolume} and integrates iterative inference using a deep temporal network. Zheng \textit{et al}.\cite{pointodyssey} improved upon PIPs by using an 8-block 1D ResNet \cite{resnet} for convolution in the temporal domain, addressing PIPs' limitations in the time dimension and achieving longer-range tracking. CoTracker \cite{cotracker} tracks multiple points jointly, accounts for their correlation, and employs a transformer model for iterative inference. TAPIR \cite{tapir} follows a coarse-to-fine tracking strategy, combining inter-frame local matching and local-correlation-based trajectory refinements. BootsTAP \cite{BootsTAP} pre-trains a "teacher" model on labeled synthetic data and uses it to generate pseudo-ground-truth labels to instruct "student" models in making predictions on unlabeled real-world data.
 
\subsection{Point Tracking Datasets}

Before the point tracking task was clarified, most video datasets with point coordinate annotations were created for specific downstream tasks. 300VW\cite{300VW} introduced facial landmark tracking data, CroHD\cite{CroHD} annotated head trajectories of pedestrians in crowds, and BADJA\cite{BADJA} marked numerous keypoints on animal joints. The specificity of video scenes and annotation points in these datasets makes it impossible to train and evaluate a universal point tracking model.
The emergence of large-scale synthetic datasets, such as those mentioned in \cite{flying, kubric, pips, autoflow}, has facilitated the pre-training of models on extensive data. Point tracking models pre-trained on these datasets demonstrate a certain degree of generalization ability when applied to real data. To further assess model performance on unseen domains, TAP-Vid \cite{TAPVid} created a challenging test set that combines real and synthetic videos from three diverse datasets \cite{kinetics, davis2017, RGBStacking}. TAP-VID\cite{TAPVid} also provide an evaluation pipeline where models are trained on a synthetic dataset \cite{kubric} and evaluated on a mixed test set.
PointOdyssey \cite{pointodyssey} utilizing real motion capture data and camera trajectories is a more complex and diverse long-term fine-grained tracking synthetic dataset, featuring an average of 18,700 trajectories and an average video length of 2,035 frames. However, the annotated points of TAP-VID \cite{TAPVid} and PointOdyssey \cite{pointodyssey} are randomly set on the surface of random entities without distinguishing the entities' texture intensity.


\section{GIFT dataset}

GIFT is a synthetic dataset designed for point-tracking tasks in indoor environments. It consists of six subsets, each formed by combining three levels of texture intensity (low, medium, high) with two types of camera trajectories (complex and normal). Each subset contains 300 videos with a resolution of 512×512, 60 frames in length, and spanning 30 different scenes. All target objects used in GIFT come from \cite{shapenet}. To enrich the video scenes, we randomly included some 3D models from GSO\cite{gso}, an open-source collection of over one thousand 3D-scanned household items, for decoration in the synthesized scenes. Additionally, HDR images collected from Poly-Haven\cite{poly-haven} were used as the background and environment for the scenes. Thanks to Kubric\cite{kubric}'s well-constructed architecture, GIFT includes rich annotations such as depth maps, instance masks, optical flow, and surface normals for each video. The uniqueness of our dataset lies in the fact that all ground truth points are set on the surfaces of classified 3D objects, ensuring that each video corresponds to a specific texture intensity.





\subsection{Texture intensity evaluation metrics}
\begin{figure}
    \centering
    \includegraphics[width=\textwidth]{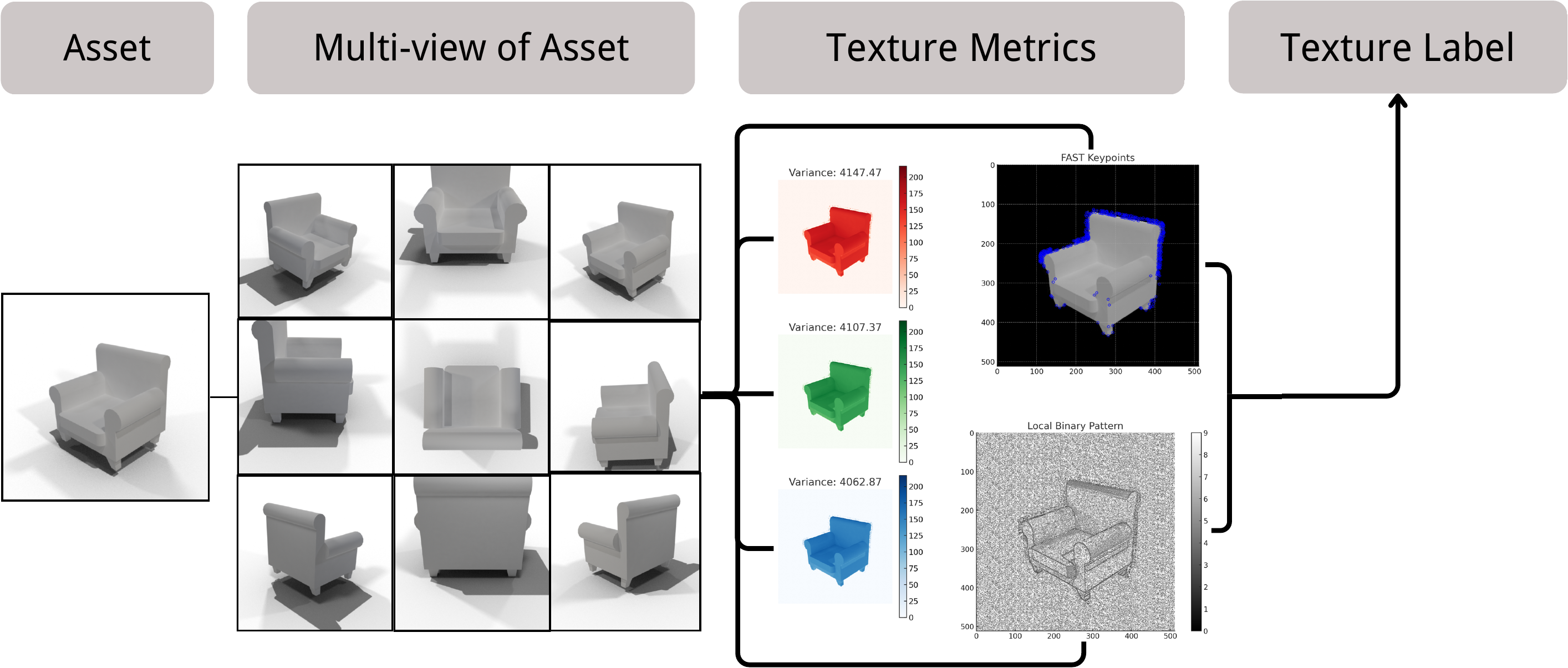}
    \caption{Categorization of asset's texture intensity level. }
    \label{fig:2}
\end{figure}
How to quantitatively evaluate the texture intensity of a 3D entity surface is a relatively unexplored problem. We simplify the problem by rendering images from different perspectives of the 3D model and then applying 2D texture evaluation metrics to these images, as shown in Figure \ref{fig:2}. This approach allows us to leverage established 2D analysis techniques while capturing the essential characteristics of the 3D model's surface texture. Specifically, we apply the following four metrics to evaluate the intensity of the texture of the rendered image. 

\subsubsection{Gray Level Co-occurrence Matrix (GLCM)}

The GLCM is used to describe the spatial distribution of pixel intensities. By dividing the image into grids of varying sizes (e.g., 5x5, 7x7, 9x9, and 11x11) centered around a query point, we calculate the following statistical features:
\begin{itemize}
    \item $Homogeneity = \sum_{i,j} \frac{P(i,j)}{1 + |i-j|}$
    \item $Energy = \sqrt{\sum_{i,j} P(i,j)^2}$
    \item $Correlation = \frac{\sum_{i,j} (i-\mu_i)(j-\mu_j)P(i,j)}{\sigma_i \sigma_j}$
    \item $Contrast = \sum_{i,j} |i-j|^2 P(i,j)$
    \item $Dissimilarity = \sum_{i,j} |i-j| P(i,j)$
\end{itemize}
where $P(i,j)$ represents the normalized GLCM matrix, and $\mu$ and $\sigma$ are the mean and standard deviation of pixel intensities.

\subsubsection{Local Binary Patterns (LBP)}

\textbf{LBP features}\cite{LBP} defined the grayscale value variation of each local position in the image. We calculate the LBP\cite{LBP} value of every pixel by
\begin{equation}
\begin{aligned}
LBP_{P,R} &= \sum_{p=0}^{P-1} s(g_p - g_0) 2^p, \quad
    s(x) = 
\begin{cases} 
1, & x \geq 0 \\
0, & x < 0 
\end{cases}
\end{aligned}
\end{equation}
where $p$ is the index of the center point and surrounding points lying on the circumference with radius $R$. The index $p$ ranges from $0$ to $P-1$, where $P$ represents the total number of sampling points on the circumference. The function $s(x)$ is a step function. In this context, $g_p$ denotes the gray-scale value of the pixel at position $p$, and $g_0$ is the gray-scale value of the center pixel. The LBP\cite{LBP} value is calculated by comparing each surrounding pixel's value $g_p$ with the center pixel's value $g_0$ and encoding the result into a binary number. This binary number is then converted to a decimal value, which serves as the LBP \cite{LBP} feature descriptor for the center pixel.
We compile the LBP\cite{LBP} features into a histogram and calculate its entropy and variance of the histogram which describe the distribution and diversity of local texture pattern. They are quantified as:
\begin{itemize}
    \item $Entropy = -\sum_{k} p_k \log_2(p_k)$, where $p_k$ is the probability of the $k$-th LBP value.
    \item $Variance = \frac{1}{N} \sum_{k=1}^N (v_k - \mu)^2$, where $v_k$ is the $k$-th LBP value, $\mu$ is the mean, and $N$ is the total number of values.
\end{itemize} 
Higher entropy means a more balanced distribution of different local texture patterns, indicating more noticeable changes in the texture of an object. Low LBP\cite{LBP} entropy represents areas with similar textures or local patterns, suggesting that their texture is uniform or repetitive. When viewed from different angles, repetitive textures remain consistent and predictable, showing uniformity from different aspects. Therefore, repetitive patterns can be regarded as uniform textures at a more macroscopic scale, and uniformity and repetitiveness are two important indicators of areas that are weakly textured.
Lower LBP\cite{LBP} variance suggests that the image textures do not vary much across the image, pointing to a uniform or smooth texture without sharp changes or distinctive edges.
The LBP method captures local texture patterns by analyzing binary differences between a pixel and its neighbors.

\subsubsection{ORB Keypoint Detection}

\textbf{The number of FAST\cite{fast} keypoints} is an important metric evaluating the level of texture intensity, FAST\cite{fast} keypoints always appear on the corners of edges or regions covered with sharp and angular patterns where points are easier to track due to their prominent visual features compared to neighboring pixels. Thus we infer that an image with less FAST\cite{fast} key points is more likely to contain weakly textured regions, and that number of key points detected in a local grid provides a measure of texture intensity:
\begin{equation}
\text{ORB Points Count} = \sum_{k=1}^M \mathbb{I}(x_k, y_k \in \text{grid})
\end{equation}
where $M$ is the total number of keypoints, and $\mathbb{I}$ is an indicator function that checks if the keypoint coordinates $(x_k, y_k)$ fall within the grid.

\subsubsection{RGB Variance Analysis}

\textbf{RGB variance} is another comprehensive metric of the degree of overall color change in an image. Lower RGB variance indicates less color change and more uniform overall color distribution in an image, which is the commonality of weakly textured regions. Texture intensity is evaluated by calculating the variance of R, G, and B channels:
\begin{itemize}
    \item Red Channel Variance: $\text{Var}_R = \frac{1}{N} \sum_{k=1}^N (R_k - \mu_R)^2$
    \item Green Channel Variance: $\text{Var}_G = \frac{1}{N} \sum_{k=1}^N (G_k - \mu_G)^2$
    \item Blue Channel Variance: $\text{Var}_B = \frac{1}{N} \sum_{k=1}^N (B_k - \mu_B)^2$
\end{itemize}
where $R_k$, $G_k$, $B_k$ are the pixel values in the respective channels, and $\mu_R$, $\mu_G$, $\mu_B$ are their means.

\subsubsection{Fourier Transform Energy}

\textbf{Low-high frequency} describe an image's texture intensity in the frequency domain. The low-frequency areas in an image often correspond to slow spatial changes, few details, and large areas of color blocks or shapes which are also the common character of the texture-less areas. More low-frequency areas means more texture-less regions while more high-frequency areas shows diverse texture patterns. The low-high frequency energy is computed as:
\begin{itemize}
    \item Low Frequency Energy: $E_{low} = \sum_{u,v \in \text{low}} |F(u,v)|^2$
    \item High Frequency Energy: $E_{high} = \sum_{u,v \notin \text{low}} |F(u,v)|^2$
\end{itemize}
where $F(u,v)$ is the Fourier transform of the grid, and "low" indicates the low-frequency components.

By integrating these metrics, we achieve a robust evaluation framework for texture intensity in 3D model surface analysis. These methods combine spatial, frequency, and color-based analyses to offer a comprehensive understanding of surface characteristics.

\begin{figure}
    \centering
    \includegraphics[scale=0.26, width=\textwidth]{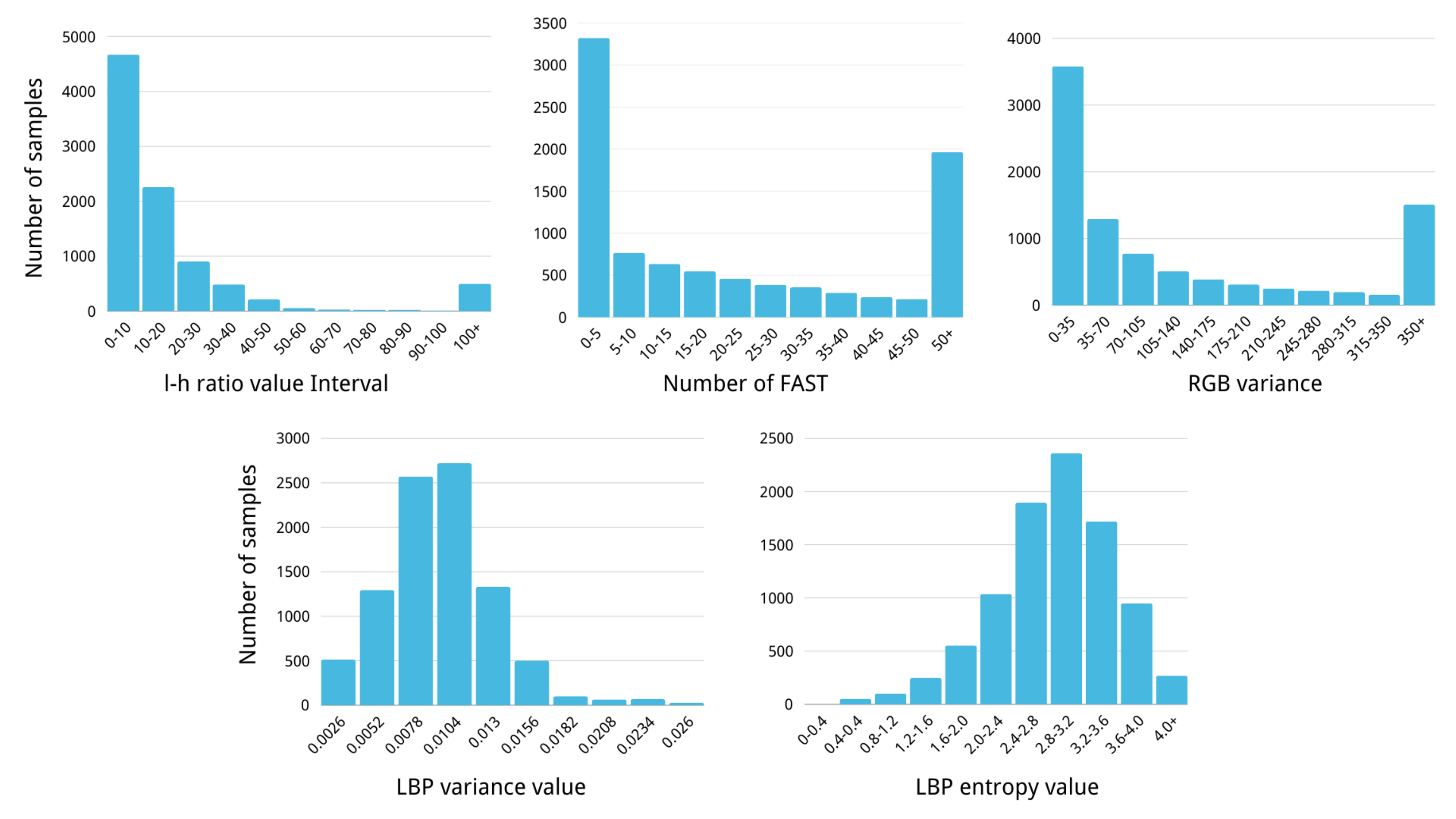}
    \caption{Statistic of frequency of 5 metrics for texture intensity evaluation.}
    \label{fig:texture-metrics}
\end{figure}

\begin{figure}[t]
    \centering
    \centerline{\includegraphics[width=.8\textwidth]{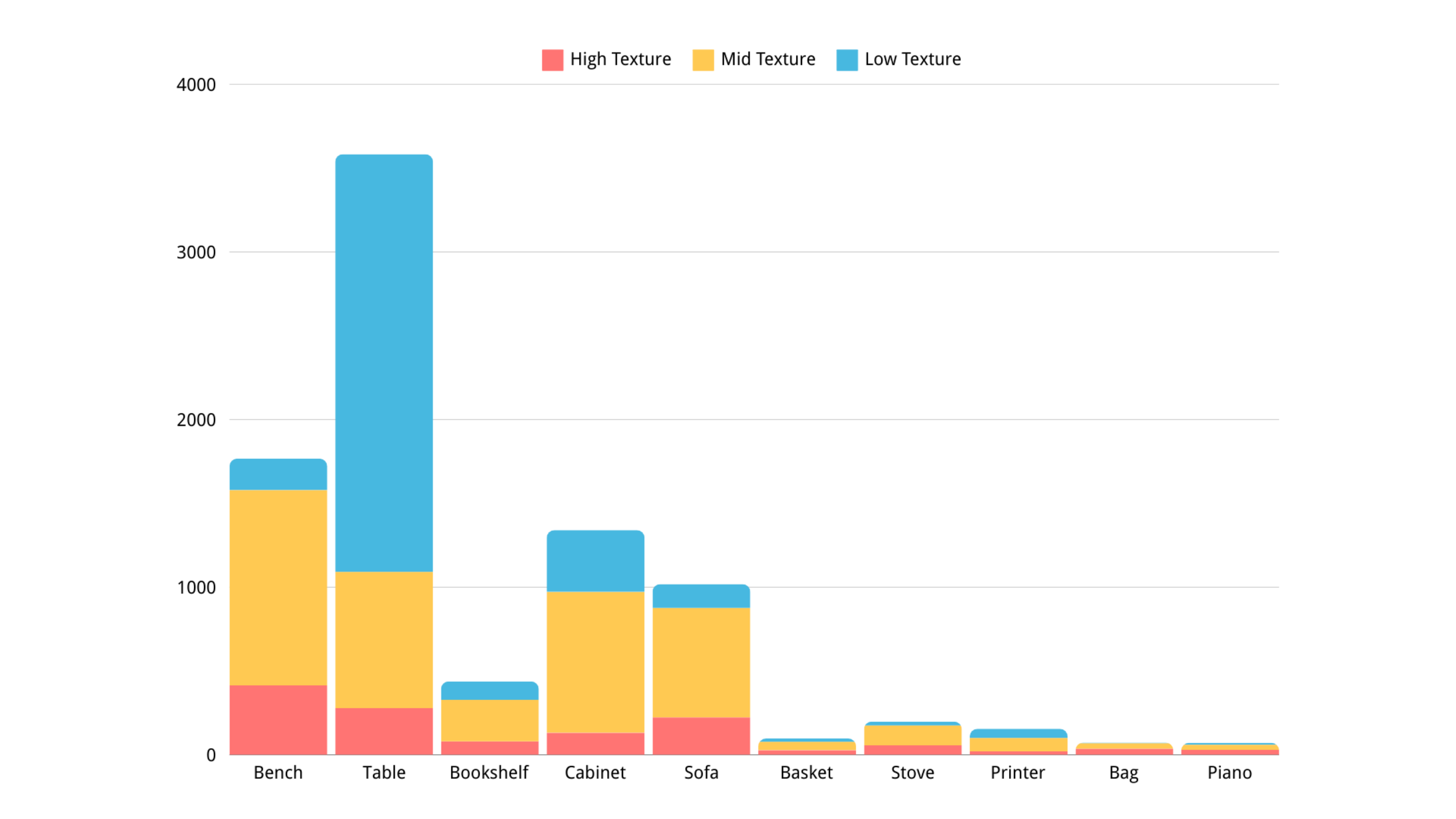}}  
    \caption{Statistics on the number of 3D models with different texture intensities for each category after classification.}
    \label{fig:category}
\end{figure}

To evaluate the relative texture intensity of a 3D object, we will calculate the average of the corresponding metrics in multi-view rendering. After obtaining the frequency histograms corresponding to each metric, we will divide it into three relative degrees in a 3:4:3 ratio. That is, every object will get a label among low, mid, and strong on each metric. We decide the final label of an entity by rating. The final texture intensity label for each object will be equivalent to the label it receives the most in each metric. If a model obtains two labels of the same quantity, we will take the label with the stronger texture. We first calculate the above-mentioned metrics separately, and the distribution of each metric is shown in Figure \ref{fig:texture-metrics}. Then we classify the objects in ShapeNet \cite{shapenet} as our source material. The statistics for each category are shown in the Figure \ref{fig:category}.  Our dataset contains mainly 10 categories of large furniture assets. Among different categories, mid-texture assets usually play dominant roles. While in the Table category, the low texture intensity becomes the major part.

\subsection{Dataset Construction}

\begin{figure}[t]
    \centering
    \includegraphics[scale=0.4]{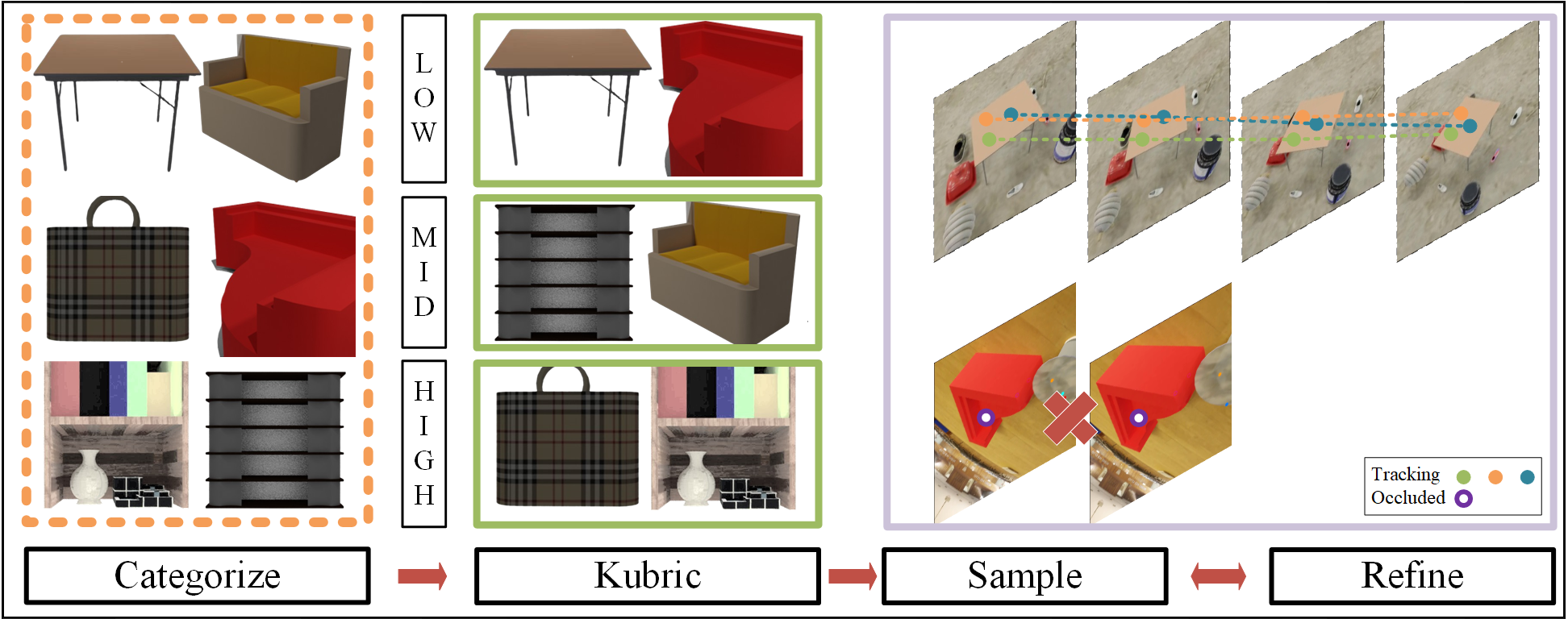}
    \caption{First, we calculate the texture intensity of 3D objects and classify them. Then, we select objects with a specific texture intensity and use Kubric\cite{kubric} to synthesize raw videos. Finally, we sample from the original videos and refine them to ensure that there are 5 to 10 visible query points.}
    \label{pipeline}
\end{figure}

After classifying the target object, our data synthesis process is divided into the following four stages: asset selection, camera trajectory generation, point annotation, and refinement. Figure \ref{pipeline} shows the pipeline for the construction of the dataset. 


\textbf{Asset selection.}
we randomly select several models with consistent texture labels as foreground objects. Ground truth points are sampled exclusively from these objects to ensure that each video corresponds to a single texture intensity level. We organize them into pre-set positions and arrange them within proper scenes. Inspired by Kubric\cite{kubric}, we select indoor backgrounds for our indoor scenes from POLYHAVEN HDRIs\cite{poly-haven} to simulate real-world scenarios.

\textbf{Camera trajectory generation.}
In order to make the videos appear more natural, we included some common camera motion strategies, such as Center Rotation (rotation around the target object), Zoom In, Zoom Out, and Linear Scanning (smooth shift of the camera's focus between different targets). Before the rendering process begins, we select different camera strategies and input them into a scheduler. The scheduler then organizes the camera's motion and introduces randomness by shuffling the sequence of strategies. It's worth noting that our camera strategies are based on human cameraman's techniques, and to ensure diverse camera dynamics across different scenes, we have incorporated random shaking into some of our videos to avoid highly similar camera motions.

\textbf{Point annotation.}
When creating the video, we specify the first object or a group of objects to be sampled. We then use the point tracking algorithm from Kubric\cite{kubric} to sample a total of 10 points. These points are randomly sampled from a specified 3D object and then mapped into the scene, and assess their visibility through depth maps and camera intrinsics and extrinsics. In the case of normal camera motion, we randomly sample points from the visible surface of the first object in the first frame. However, for complex camera motion, we take the first set of objects and randomly sample points from them when they are visible. Normally, the first frame has 5 to 10 visible points under both two camera motion strategies.

\textbf{Refinement.}
To ensure the quality of our videos, we carefully inspect each sample in our dataset for various flaws and defects. Since kubric's algorithm estimates the position of points through mathematical formulas, there is a certain error in the estimation of the visibility of points. These include incorrectly annotated invisible target points, target points that do not align with the surface of the target objects, and samples where most of the targets are obscured.  We conduct two rounds of quality checks for video frames and point annotations. First, when 3D scenes are converted into RGB frames, a human annotator will review the visual results without point annotations to identify any samples with visual flaws. Then, in stage 3, a different annotator will conduct a second check of the visual results, To ensure enough visible points in the video, we use the "queried first" protocol, which means that each point is queried only once in the video at the first frame where it becomes visible. If more than 5 points are not visible throughout the video, the video will be removed. This process ensures that valid sampled points in the video range from 5 to 10.

\section{Benchmark and Baseline}
\subsection{Evaluation on the datsets}
\subsubsection{Evaluation Results of Different Datasets}
\begin{figure}
    \centering
    \includegraphics[scale=1.0, width=\textwidth]{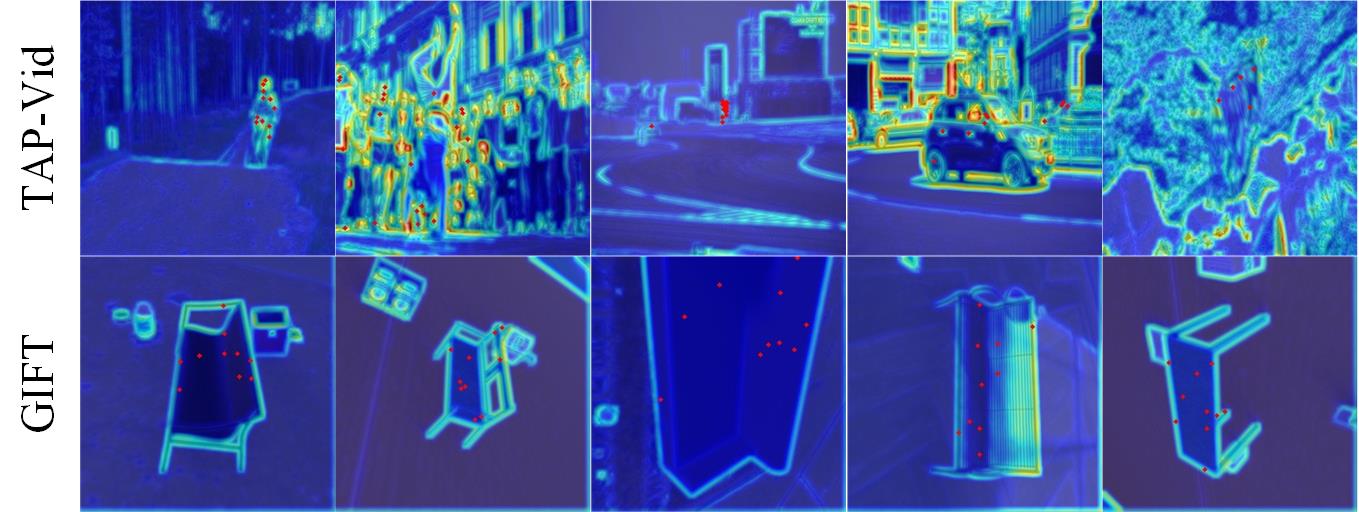}
    \caption{Based on the visualization results of RGB Variance, the redder the region is, the stronger the texture strength is, and the bluer the region is, the weaker the texture is. From the figure, we can see that the texture intensity of TapVid\cite{TAPVid} is stronger compared to our dataset\label{fig:resultVis1}}
\end{figure}

To prove the rationality of our proposed texture indicators, we used the texture evaluation formula and performed the texture evaluation on different test machines. By evaluating the average texture intensity of each video at a point, the average texture intensity of the entire dataset is calculated. Table \ref{datasetCompare} compares our dataset with \cite{TAPVid} and Figure \ref{fig:resultVis1} is the visualized result of the texture analysis using our texture indicators.
\begin{table}[h!]
\centering
\begin{tabular}{c c c}
\hline
\multirow{2}{*}{\textbf{Metrics}} & \textbf{TAP-Vid-Davis\cite{TAPVid}} & \textbf{GIFT(N-L)} \\ \cmidrule(r){2-2}\cmidrule(r){3-3}
                              & Value              & Value             \\ \hline
\textbf{Corner Points} & \multicolumn{2}{c}{} \\ 
orb\_points                    & 0.701 & 0.743 \\ \hline
\textbf{Fourier Transform} & \multicolumn{2}{c}{} \\ 
high\_freq\_energy             & 437  & 35.5 \\
low\_freq\_energy              & 463000  & 1170000 \\\hline
\textbf{GLCM} & \multicolumn{2}{c}{} \\
contrast                       & 10700 & 7450 \\ 
correlation                    & 0.00860 & 0.112 \\ 
dissimilarity                  & 82.8   & 58.7  \\ 
energy                         & 0.108 & 0.190 \\ 
homogeneity                    & 0.020 & 0.220 \\ \hline
\textbf{LBP} & \multicolumn{2}{c}{} \\ 
lbp\_entropy                   & 3.16   & 3.17 \\ 
lbp\_variance                  & 85.2 & 76.8  \\ \hline
\textbf{RGB Variance} & \multicolumn{2}{c}{} \\ 
b\_var                         & 1450 & 193 \\ 
g\_var                         & 1400 & 167 \\ 
r\_var                         & 1500 & 163 \\ \hline
\end{tabular}
\caption{Comparion between two datasets\label{datasetCompare}}
\end{table}
\subsection{Evaluation on the model}
\subsubsection{Evaluation Metrics}
We adopt the same metrics as TAP-Vid\cite{TAPVid} to evaluate the model: Occlusion Accuracy (OA) measures the precision of point occlusion prediction on each frame. $\delta_{avg}^{vis}$ measures the average fraction of visible points that fall within each threshold of their ground truth. The thresholds are set at 1, 2, 4, 8, and 16 pixels.
Average Jaccard measures both occlusion and position accuracy.

\subsubsection{Evaluation Results of Different Datasets}
In the evaluation, we employed six models, which have performed well on datasets like TAP-Vid-Davis\cite{TAPVid}, to assess our textureless dataset. For evaluating RAFT\cite{raft}, we chain the flow. When the travel is out of bounds, we clamp the coordinates to the image bounds and sample at the edge of the flow map. We adopted the strategy of "queried first" to evaluate our dataset. Table \ref{tab:evaluation_results} displays the experimental results. Results illustrate that all models have difficulty in tracking in textureless areas.

\begin{table}[htbp]
    \centering
    \caption{Evaluation Results}
    \label{tab:evaluation_results}
    \begin{tabular}{c c c c c c c}
        \toprule
        \multirow{1}{*}{Model} & \multicolumn{3}{c}{TAP-Vid-Davis\cite{TAPVid}}  &\multicolumn{3}{c}{GIFT(N-L)}\\ 
        \cmidrule(r){2-4} \cmidrule(r){5-7} & $AJ\uparrow$ & $\delta_{avg}^{vis}\uparrow$  & $OA\uparrow$ & $AJ\uparrow$ & $\delta_{avg}^{vis}\uparrow$  & $OA\uparrow$ \\ \bottomrule
        RAFT\cite{raft} & -& 35.9& -& -& 37.6& -\\
        TAP-Net\cite{TAPVid} & 32.7&48.4 & 77.6& 27.3& 39.1& 85.7\\
        PIPs++\cite{pointodyssey}& -& 62.9& -& -& 50.0& - \\ 
        TAPIR\cite{tapir}& 58.5& 70.6& 87.3& 41.6& 52.1& 75.8\\ 
        CoTracker\cite{cotracker}& 60.9& 75.4& 88.4& 43.4& 52.0& 83.3\\ 
        BootsTAP\cite{BootsTAP}& 61.1& 73.6& 88.2& 46.5& 56.2& 84.2\\ \bottomrule
    \end{tabular}
\end{table}
\subsubsection{Evaluation between Different Levels}
We synthesized six distinct datasets, each varying according to the levels of camera motion and object texture intensity. The camera motion was categorized into two levels: \textbf{normal} and \textbf{complex}. The texture intensity was divided into three levels: \textbf{low}, \textbf{medium}, and \textbf{high}. Table \ref{tab:evaluation_on_diff} shows the result that we tested our six levels of data sets on different models.

\begin{table}[t]
    \centering
    \caption{Experiments on different levels.}
    \label{tab:evaluation_on_diff}
    \resizebox{\textwidth}{!}{
        \begin{tabular}{c c c c c c c c c c c c c c c c c c c}
        \toprule
        Pattern(Camera-Texture) & \multicolumn{3}{c}{N-L}& \multicolumn{3}{c}{N-M} & \multicolumn{3}{c}{N-H} &\multicolumn{3}{c}{C-L} & 
        \multicolumn{3}{c}{C-M} & \multicolumn{3}{c}{C-H} \\
        \cmidrule(r){2-4} \cmidrule(r){5-7} \cmidrule(r){8-10} \cmidrule(r){11-13} \cmidrule(r){14-16} \cmidrule(r){17-19}
        Model & $AJ\uparrow$ & $\delta_{avg}^{vis}\uparrow$  & $OA\uparrow$ & $AJ\uparrow$ & $\delta_{avg}^{vis}\uparrow$  & $OA\uparrow$ & $AJ\uparrow$ & $\delta_{avg}^{vis}\uparrow$  & $OA\uparrow$ & $AJ\uparrow$ & $\delta_{avg}^{vis}\uparrow$  & $OA\uparrow$ & $AJ\uparrow$ & $\delta_{avg}^{vis}\uparrow$  & $OA\uparrow$ & $AJ\uparrow$ & $\delta_{avg}^{vis}\uparrow$  & $OA\uparrow$  \\
        \bottomrule
        RAFT\cite{raft} & -& 37.6& -& -& 40.5& -& -& 45.1& -& -& 26.7& -& -& 30.8& -& -& 33.6& -\\
        TAPNet\cite{TAPVid} & 27.3& 39.1& 85.7& 30.9& 43.4& 87.0& 27.3& 41.1& 80.9& 28.0& 42.5& 77.9& 31.0& 46.6& 77.4& 34.0& 49.6& 78.5\\
        PIPs++\cite{pointodyssey} & -& 50.0& -& -& 54.3& -& -& 58.9& -& -& 35.8& -& -& 41.1& -& -& 44.9& - \\
        TAPIR\cite{tapir} & 41.6& 52.1& 75.8& 46.6& 57.8& 81.3& 50.2& 62.0& 83.2& 46.5& 57.0& 80.3& 52.4& 62.6& 84.7& 56.7& 67.0& 85.8 \\
        CoTracker\cite{cotracker}  & 43.4& 52.0& 83.3& 48.7& 58.9& 86.0& 58.2& 69.9& 88.0& 30.1& 37.4& 72.8& 37.8& 45.3& 75.0& 41.5& 48.8& 75.8\\
        BootsTAP\cite{BootsTAP} & 46.5& 56.2& 84.2& 52.1& 63.0& 86.9& 56.6& 69.0&86.3 & 50.0& 61.3& 82.9& 56.9& 68.2& 86.9& 61.5& 72.0& 87.5\\
        \bottomrule
        \end{tabular}
    }

    \label{tab:experment2}

\end{table}
\subsection{Analysis}
Our experiments revealed that certain baseline algorithms demonstrated reduced accuracy as the complexity of camera motion increased. Surprisingly, TAPIR\cite{tapir} and BootsTAPIR\cite{BootsTAP} exhibited improved performance compared to their performance on previous subsets with normal camera motions. This phenomenon is attributed to the robust nature of TAPIR's\cite{tapir} global search in handling occlusions. In our dataset, the subsets with complex camera motions largely entailed camera focus shifting, resulting in continuous changes in the objects captured in the videos. Within this context, models' capability to handle heavy occlusions dominates their prediction accuracy.

\begin{figure}
    \centering
    \includegraphics[scale=1.0, width=\textwidth]{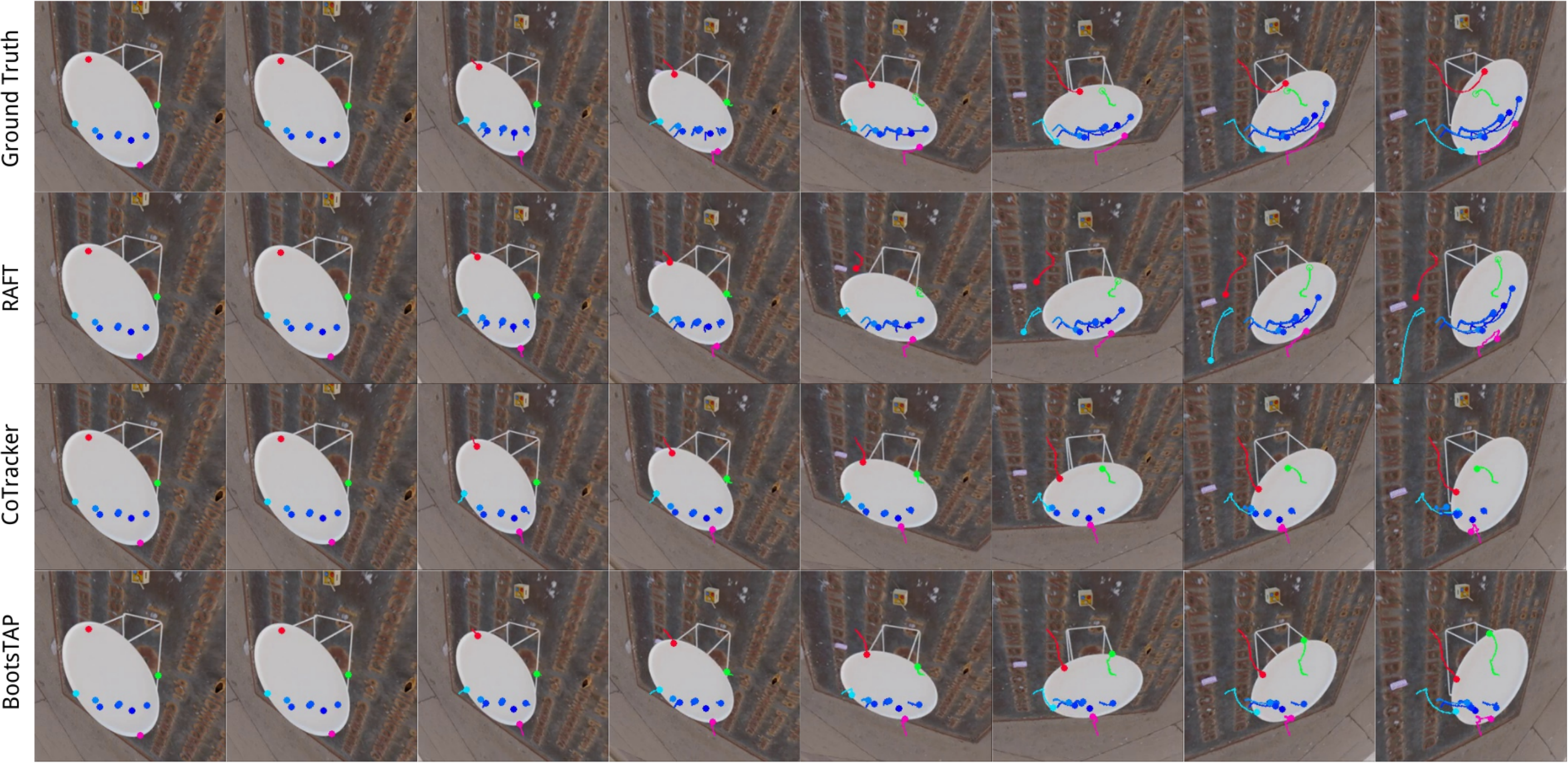}
    \caption{The comparative visualization of RAFT\cite{raft}, CoTracker\cite{cotracker}, and BootsTAP\cite{BootsTAP} performance indicates distinct strengths and weaknesses in point tracking. RAFT\cite{raft} effectively predicts the motion trajectory within textureless regions; however, it struggles with maintaining edge points, leading to their loss. Conversely, both CoTracker\cite{cotracker} and BootsTAP\cite{BootsTAP} exhibit difficulty in tracking within textureless regions, resulting in the loss of tracking points in these areas.}
    \label{fig:resultVis2}
\end{figure}

Algorithms such as CoTracker\cite{cotracker} and PIPs\cite{pips} relied on local neighborhood search and temporal track smoothing. However, their sequential video processing in chunks, with each chunk initialized using the previous output, led to challenges in handling occlusions. In contrast, the TAP-Net series\cite{tapir}\cite{BootsTAP}\cite{TAPVid} adopted a global search approach, conducting an independent global search for query points on each frame, thereby establishing coarse tracks that are more robust against occlusions. We visualize the tracking result of the above-mentioned algorithms on a weakly textured video track in our dataset, as shown in Figure \ref{fig:resultVis2}.

Ideas and strategies can also be used to address the challenges of tracking weakly-textured areas. Many previous works have focused on smoothing the tracks and improving the accuracy of matching neighboring pixels between frames. However, they often fail to fully utilize information from a global perspective. When target points fall into poorly textured regions, existing algorithms can easily be confused by uniform and repetitive local features, leading to a decrease in tracking accuracy. With this texture-less context, a global search similar to TAPIR can be efficient.

\section{Discussion}
\subsection{Limitations and future work}
While our human annotators have carefully inspected GIFT's building process, there are still potential risks for annotators to make mistakes since checking annotation points on poorly textured areas can be difficult and exhausting. Meanwhile, the camera motion strategies in GIFT are not yet fully developed. Numerous combinations of camera techniques do not accurately replicate the smoothness of human camerawork.

\subsection{Potential negative societal impacts}
Constructing synthetic datasets is crucial for engineers to identify and rectify potential failure modes in applications involving human interaction, like self-driving cars and robotics. This proactive approach reduces the risk of harm to humans before real-world deployment. However, it's imperative to recognize the possibility of developing systems that could cause serious injury. We will always prioritize safety in the design process. and explore the proper usage of synthetic datasets to prevent harmful outcomes is essential.

\section{Conclusion}

In this paper, we introduce GIFT, an indoor dataset for point tracking tasks over poorly textured regions. We  first design combined metrics for quantitative evaluation  of assets' texture intensity. Then we utilize Kubric, combined with carefully selected furniture with different texture labels and background materials, to provide a testing dataset for the evaluation of point-tracking algorithms. With this approach, we not only provide a series of challenging tracking tasks over texture-free regions but also give insights on possible improvements for future works. The diversity and detailed annotation of the dataset are one of its core strengths, and they provide researchers with a wealth of information to better understand and solve complex problems in point-tracking processes. 


However, our dataset still has some limitations in modeling real-world scenarios. Specifically, we need to expand the size of the dataset and include a greater diversity of asset categories and scene types to cover a wider range of real-world application scenarios. Additionally, since our data comes from existing public sources, researchers should use it carefully and be mindful of potential biases within the data source.

At the same time, we have reason to believe that GIFT will play an even more important role in future scientific research and technological development, especially in advancing autonomous navigation and augmented reality technologies. We expect the dataset to stimulate new research ideas, promote interdisciplinary collaboration, and provide innovative solutions to practical problems.
\small
\bibliographystyle{unsrt}

\begin{thebibliography}{10}

\bibitem{shapenet}
Angel~X Chang, Thomas Funkhouser, Leonidas Guibas, Pat Hanrahan, Qixing Huang, Zimo Li, Silvio Savarese, Manolis Savva, Shuran Song, Hao Su, et~al.
\newblock Shapenet: An information-rich 3d model repository.
\newblock {\em arXiv preprint arXiv:1512.03012}, 2015.

\bibitem{sam-pt}
Frano Raji{\v{c}}, Lei Ke, Yu-Wing Tai, Chi-Keung Tang, Martin Danelljan, and Fisher Yu.
\newblock Segment anything meets point tracking.
\newblock {\em arXiv preprint arXiv:2307.01197}, 2023.

\bibitem{surgical-sam}
Zijian Wu, Adam Schmidt, Peter Kazanzides, and Septimiu~E Salcudean.
\newblock Real-time surgical instrument segmentation in video using point tracking and segment anything.
\newblock {\em arXiv preprint arXiv:2403.08003}, 2024.

\bibitem{videoswap}
Yuchao Gu, Yipin Zhou, Bichen Wu, Licheng Yu, Jia-Wei Liu, Rui Zhao, Jay~Zhangjie Wu, David~Junhao Zhang, Mike~Zheng Shou, and Kevin Tang.
\newblock Videoswap: Customized video subject swapping with interactive semantic point correspondence.
\newblock {\em arXiv preprint arXiv:2312.02087}, 2023.

\bibitem{videoshop}
Xiang Fan, Anand Bhattad, and Ranjay Krishna.
\newblock Videoshop: Localized semantic video editing with noise-extrapolated diffusion inversion.
\newblock {\em arXiv preprint arXiv:2403.14617}, 2024.

\bibitem{videoMotion}
Zeqi Xiao, Yifan Zhou, Shuai Yang, and Xingang Pan.
\newblock Video diffusion models are training-free motion interpreter and controller.
\newblock {\em arXiv preprint arXiv:2405.14864}, 2024.

\bibitem{revideo}
Chong Mou, Mingdeng Cao, Xintao Wang, Zhaoyang Zhang, Ying Shan, and Jian Zhang.
\newblock Revideo: Remake a video with motion and content control.
\newblock {\em arXiv preprint arXiv:2405.13865}, 2024.

\bibitem{mft}
Michal Neoral, Jon{\'a}{\v{s}} {\v{S}}er{\`y}ch, and Ji{\v{r}}{\'\i} Matas.
\newblock Mft: Long-term tracking of every pixel.
\newblock In {\em Proceedings of the IEEE/CVF Winter Conference on Applications of Computer Vision}, pages 6837--6847, 2024.

\bibitem{advert3D}
Ivan Bacher, Hossein Javidnia, Soumyabrata Dev, Rahul Agrahari, Murhaf Hossari, Matthew Nicholson, Clare Conran, Jian Tang, Peng Song, David Corrigan, et~al.
\newblock An advert creation system for 3d product placements.
\newblock In {\em Machine Learning and Knowledge Discovery in Databases: Applied Data Science Track: European Conference, ECML PKDD 2020, Ghent, Belgium, September 14--18, 2020, Proceedings, Part IV}, pages 224--239. Springer, 2021.

\bibitem{motionDrag}
Ruining Li, Chuanxia Zheng, Christian Rupprecht, and Andrea Vedaldi.
\newblock Dragapart: Learning a part-level motion prior for articulated objects.
\newblock {\em arXiv preprint arXiv:2403.15382}, 2024.

\bibitem{motionI2V}
Xiaoyu Shi, Zhaoyang Huang, Fu-Yun Wang, Weikang Bian, Dasong Li, Yi~Zhang, Manyuan Zhang, Ka~Chun Cheung, Simon See, Hongwei Qin, et~al.
\newblock Motion-i2v: Consistent and controllable image-to-video generation with explicit motion modeling.
\newblock {\em arXiv preprint arXiv:2401.15977}, 2024.

\bibitem{motionIpose}
Jingyuan Liu, Li-Yi Wei, Ariel Shamir, and Takeo Igarashi.
\newblock ipose: Interactive human pose reconstruction from video.
\newblock In {\em Proceedings of the CHI Conference on Human Factors in Computing Systems}, pages 1--14, 2024.

\bibitem{TAPVid}
Carl Doersch, Ankush Gupta, Larisa Markeeva, Adri{\`a} Recasens, Lucas Smaira, Yusuf Aytar, Jo{\~a}o Carreira, Andrew Zisserman, and Yezhou Yang.
\newblock Tap-vid: A benchmark for tracking any point in a video.
\newblock {\em ArXiv}, abs/2211.03726, 2022.

\bibitem{pips}
Adam~W Harley, Zhaoyuan Fang, and Katerina Fragkiadaki.
\newblock Particle video revisited: Tracking through occlusions using point trajectories.
\newblock In {\em European Conference on Computer Vision}, pages 59--75. Springer, 2022.

\bibitem{tapir}
Carl Doersch, Yi~Yang, Mel Vecerik, Dilara Gokay, Ankush Gupta, Yusuf Aytar, Joao Carreira, and Andrew Zisserman.
\newblock Tapir: Tracking any point with per-frame initialization and temporal refinement.
\newblock In {\em Proceedings of the IEEE/CVF International Conference on Computer Vision}, pages 10061--10072, 2023.

\bibitem{omnimotion}
Qianqian Wang, Yen-Yu Chang, Ruojin Cai, Zhengqi Li, Bharath Hariharan, Aleksander Holynski, and Noah Snavely.
\newblock Tracking everything everywhere all at once.
\newblock In {\em Proceedings of the IEEE/CVF International Conference on Computer Vision}, pages 19795--19806, 2023.

\bibitem{BootsTAP}
Carl Doersch, Yi~Yang, Dilara Gokay, Pauline Luc, Skanda Koppula, Ankush Gupta, Joseph Heyward, Ross Goroshin, Jo{\~a}o Carreira, and Andrew Zisserman.
\newblock Bootstap: Bootstrapped training for tracking-any-point.
\newblock {\em arXiv preprint arXiv:2402.00847}, 2024.

\bibitem{spatialtracker}
Yuxi Xiao, Qianqian Wang, Shangzhan Zhang, Nan Xue, Sida Peng, Yujun Shen, and Xiaowei Zhou.
\newblock Spatialtracker: Tracking any 2d pixels in 3d space.
\newblock {\em arXiv preprint arXiv:2404.04319}, 2024.

\bibitem{pointodyssey}
Yang Zheng, Adam~W Harley, Bokui Shen, Gordon Wetzstein, and Leonidas~J Guibas.
\newblock Pointodyssey: A large-scale synthetic dataset for long-term point tracking.
\newblock In {\em Proceedings of the IEEE/CVF International Conference on Computer Vision}, pages 19855--19865, 2023.

\bibitem{LBP}
Timo Ojala, Matti Pietikainen, and Topi Maenpaa.
\newblock Multiresolution gray-scale and rotation invariant texture classification with local binary patterns.
\newblock {\em IEEE Transactions on pattern analysis and machine intelligence}, 24(7):971--987, 2002.

\bibitem{fast}
Edward Rosten and Tom Drummond.
\newblock Machine learning for high-speed corner detection.
\newblock In {\em Computer Vision--ECCV 2006: 9th European Conference on Computer Vision, Graz, Austria, May 7-13, 2006. Proceedings, Part I 9}, pages 430--443. Springer, 2006.

\bibitem{kubric}
Klaus Greff, Francois Belletti, Lucas Beyer, Carl Doersch, Yilun Du, Daniel Duckworth, David~J Fleet, Dan Gnanapragasam, Florian Golemo, Charles Herrmann, et~al.
\newblock Kubric: A scalable dataset generator.
\newblock In {\em Proceedings of the IEEE/CVF conference on computer vision and pattern recognition}, pages 3749--3761, 2022.

\bibitem{MPI}
Daniel~J. Butler, Jonas Wulff, Garrett~B. Stanley, and Michael~J. Black.
\newblock A naturalistic open source movie for optical flow evaluation.
\newblock In Andrew Fitzgibbon, Svetlana Lazebnik, Pietro Perona, Yoichi Sato, and Cordelia Schmid, editors, {\em Computer Vision -- ECCV 2012}, pages 611--625, Berlin, Heidelberg, 2012. Springer Berlin Heidelberg.

\bibitem{flying}
Nikolaus Mayer, Eddy Ilg, Philip Hausser, Philipp Fischer, Daniel Cremers, Alexey Dosovitskiy, and Thomas Brox.
\newblock A large dataset to train convolutional networks for disparity, optical flow, and scene flow estimation.
\newblock In {\em Proceedings of the IEEE conference on computer vision and pattern recognition}, pages 4040--4048, 2016.

\bibitem{kinetics}
Joao Carreira and Andrew Zisserman.
\newblock Quo vadis, action recognition? a new model and the kinetics dataset.
\newblock In {\em proceedings of the IEEE Conference on Computer Vision and Pattern Recognition}, pages 6299--6308, 2017.

\bibitem{davis2017}
Jordi Pont-Tuset, Federico Perazzi, Sergi Caelles, Pablo Arbel{\'a}ez, Alex Sorkine-Hornung, and Luc Van~Gool.
\newblock The 2017 davis challenge on video object segmentation.
\newblock {\em arXiv preprint arXiv:1704.00675}, 2017.

\bibitem{flownet}
Philipp Fischer, Alexey Dosovitskiy, Eddy Ilg, Philip H{\"a}usser, Caner Haz{\i}rba{\c{s}}, Vladimir Golkov, Patrick Van~der Smagt, Daniel Cremers, and Thomas Brox.
\newblock Flownet: Learning optical flow with convolutional networks.
\newblock {\em arXiv preprint arXiv:1504.06852}, 2015.

\bibitem{raft}
Zachary Teed and Jia Deng.
\newblock Raft: Recurrent all-pairs field transforms for optical flow.
\newblock In {\em Computer Vision--ECCV 2020: 16th European Conference, Glasgow, UK, August 23--28, 2020, Proceedings, Part II 16}, pages 402--419. Springer, 2020.

\bibitem{loftr}
Jiaming Sun, Zehong Shen, Yuang Wang, Hujun Bao, and Xiaowei Zhou.
\newblock Loftr: Detector-free local feature matching with transformers.
\newblock In {\em Proceedings of the IEEE/CVF conference on computer vision and pattern recognition}, pages 8922--8931, 2021.

\bibitem{videoflow}
Xiaoyu Shi, Zhaoyang Huang, Weikang Bian, Dasong Li, Manyuan Zhang, Ka~Chun Cheung, Simon See, Hongwei Qin, Jifeng Dai, and Hongsheng Li.
\newblock Videoflow: Exploiting temporal cues for multi-frame optical flow estimation.
\newblock In {\em Proceedings of the IEEE/CVF International Conference on Computer Vision}, pages 12469--12480, 2023.

\bibitem{cotracker}
Nikita Karaev, Ignacio Rocco, Benjamin Graham, Natalia Neverova, Andrea Vedaldi, and Christian Rupprecht.
\newblock Cotracker: It is better to track together.
\newblock {\em arXiv preprint arXiv:2307.07635}, 2023.

\bibitem{track_everything}
Qianqian Wang, Yen-Yu Chang, Ruojin Cai, Zhengqi Li, Bharath Hariharan, Aleksander Holynski, and Noah Snavely.
\newblock Tracking everything everywhere all at once.
\newblock In {\em Proceedings of the IEEE/CVF International Conference on Computer Vision}, pages 19795--19806, 2023.

\bibitem{costVolume}
Asmaa Hosni, Christoph Rhemann, Michael Bleyer, Carsten Rother, and Margrit Gelautz.
\newblock Fast cost-volume filtering for visual correspondence and beyond.
\newblock {\em IEEE transactions on pattern analysis and machine intelligence}, 35(2):504--511, 2012.

\bibitem{4DCostVolume}
Deqing Sun, Xiaodong Yang, Ming-Yu Liu, and Jan Kautz.
\newblock Pwc-net: Cnns for optical flow using pyramid, warping, and cost volume.
\newblock In {\em Proceedings of the IEEE conference on computer vision and pattern recognition}, pages 8934--8943, 2018.

\bibitem{resnet}
Kaiming He, Xiangyu Zhang, Shaoqing Ren, and Jian Sun.
\newblock Deep residual learning for image recognition.
\newblock In {\em Proceedings of the IEEE conference on computer vision and pattern recognition}, pages 770--778, 2016.

\bibitem{300VW}
Jie Shen, Stefanos Zafeiriou, Grigoris~G Chrysos, Jean Kossaifi, Georgios Tzimiropoulos, and Maja Pantic.
\newblock The first facial landmark tracking in-the-wild challenge: Benchmark and results.
\newblock In {\em Proceedings of the IEEE international conference on computer vision workshops}, pages 50--58, 2015.

\bibitem{CroHD}
Ramana Sundararaman, Cedric De~Almeida~Braga, Eric Marchand, and Julien Pettre.
\newblock Tracking pedestrian heads in dense crowd.
\newblock In {\em Proceedings of the IEEE/CVF conference on computer vision and pattern recognition}, pages 3865--3875, 2021.

\bibitem{BADJA}
Benjamin Biggs, Thomas Roddick, Andrew Fitzgibbon, and Roberto Cipolla.
\newblock Creatures great and smal: Recovering the shape and motion of animals from video.
\newblock In {\em Computer Vision--ACCV 2018: 14th Asian Conference on Computer Vision, Perth, Australia, December 2--6, 2018, Revised Selected Papers, Part V 14}, pages 3--19. Springer, 2019.

\bibitem{autoflow}
Deqing Sun, Daniel Vlasic, Charles Herrmann, Varun Jampani, Michael Krainin, Huiwen Chang, Ramin Zabih, William~T Freeman, and Ce~Liu.
\newblock Autoflow: Learning a better training set for optical flow.
\newblock In {\em Proceedings of the IEEE/CVF Conference on Computer Vision and Pattern Recognition}, pages 10093--10102, 2021.

\bibitem{RGBStacking}
Alex~X Lee, Coline~Manon Devin, Yuxiang Zhou, Thomas Lampe, Konstantinos Bousmalis, Jost~Tobias Springenberg, Arunkumar Byravan, Abbas Abdolmaleki, Nimrod Gileadi, David Khosid, et~al.
\newblock Beyond pick-and-place: Tackling robotic stacking of diverse shapes.
\newblock In {\em 5th Annual Conference on Robot Learning}, 2021.

\bibitem{gso}
Laura Downs, Anthony Francis, Nate Koenig, Brandon Kinman, Ryan Hickman, Krista Reymann, Thomas~B McHugh, and Vincent Vanhoucke.
\newblock Google scanned objects: A high-quality dataset of 3d scanned household items.
\newblock In {\em 2022 International Conference on Robotics and Automation (ICRA)}, pages 2553--2560. IEEE, 2022.

\bibitem{poly-haven}
Greg, Rob, Rico, James, Andreas, Sergej, Dimitrios, and Jurita.
\newblock Polyhaven: a curated public asset library for visual effects artists and game designers.
\newblock 2021.

\bibitem{lvos}
Lingyi Hong, Zhongying Liu, Wenchao Chen, Chenzhi Tan, Yuang Feng, Xinyu Zhou, Pinxue Guo, Jinglun Li, Zhaoyu Chen, Shuyong Gao, et~al.
\newblock Lvos: A benchmark for large-scale long-term video object segmentation.
\newblock {\em arXiv preprint arXiv:2404.19326}, 2024.

\end{thebibliography}

\end{document}


\maketitle

\begin{appendix}
\section{Appendix}
\subsection{Experiment Details}
The data dimension of the GIFT is consistent with that of TAP-Vid-Davis\cite{davis2017}. We downsample all the videos to 256 x 256 (originally 512 x 512) and map the coordinates of points to [0, 255] (originally [0, 511]). We then resize the videos to the resolution they are trained at 512 x 896 for PIPs++\cite{pips}, and 384 x 512 for CoTracker\cite{cotracker}. We evaluate our dataset using hyperparameters provided by the models' github repository for the evaluation of TAP-Vid-Davis\cite{davis2017}. For RAFT\cite{raft}, we set the number of iterative updates to 24. We use TAP-Vid\cite{TAPVid}'s metrics to evaluate models' performance. We use the weight file provided by the models' GitHub repository to evaluate our dataset, and for RAFT\cite{raft}, we use raft-things.pth as the weight file. Our experimental environment uses NVIDIA A100-SXM4-40GB.

\section{GIFT Dataset}

\textbf{Terms of use, privacy and License: }
The dataset is distributed under the Creative Commons 4.0 BY-NC-SA license.

\textbf{Data statistics and maintenance: }github repository: \href{https://anonymous.4open.science/r/GIFT-6D02/}{https://anonymous.4open.science/r/GIFT-6D02/}, download url: \href{https://terabox.com/s/1sVErlKGtz0uf6OrBj4wF4g}{https://terabox.com/s/1sVErlKGtz0uf6OrBj4wF4g}

\section{Additional Analysis}
We compare the performance of CoTracker\cite{cotracker} and BootsTAP\cite{BootsTAP} on our GIFT dataset. There exists an obvious difference in these two baselines' performance of the GIFT subsets with complex camera motions. We introduce complex combinations of camera strategies, potentially introducing occlusions to many track targets in the data. Since occlusions may bring much more difficulties to point tracking algorithms, we decided to analyze the correlation between models' occlusion accuracies and point prediction accuracies.

As shown in Fig 1, BootsTAP performs over-all better on scenes among weekly textured, mid-textured, and high-textured areas. A strong correlation between occlusion accuracy and point position accuracy is shown in Fig.3. The accuracy difference on both x and y axis are calculated by BootsTAP's minus CoTracker's.

\begin{figure}
    \centering
    \includegraphics[scale=0.3]{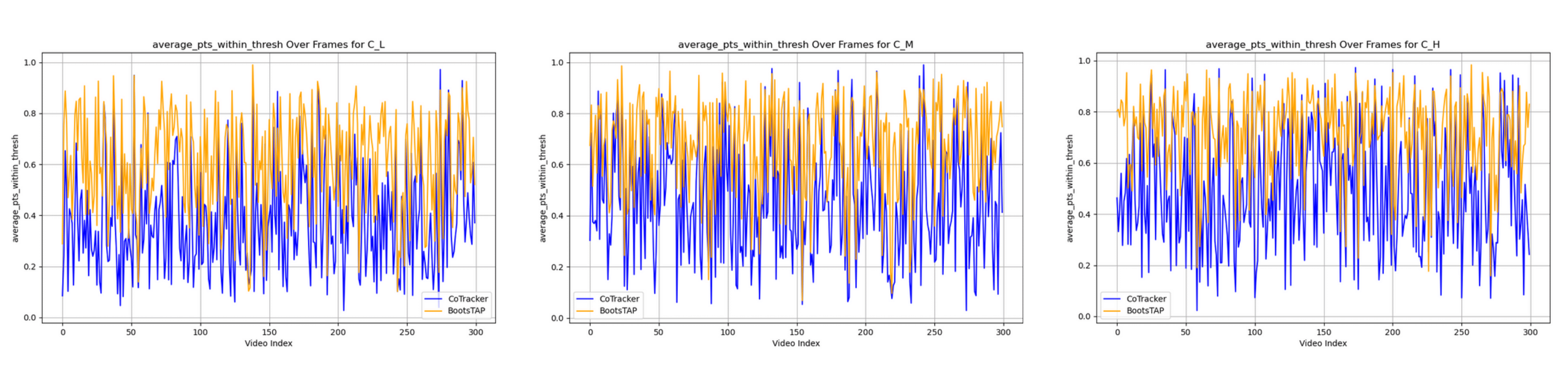}
    \caption{Comparison of point accuracy}
    \label{fig:enter-label}
\end{figure}

\begin{figure}
    \centering
    \includegraphics[scale=0.3]{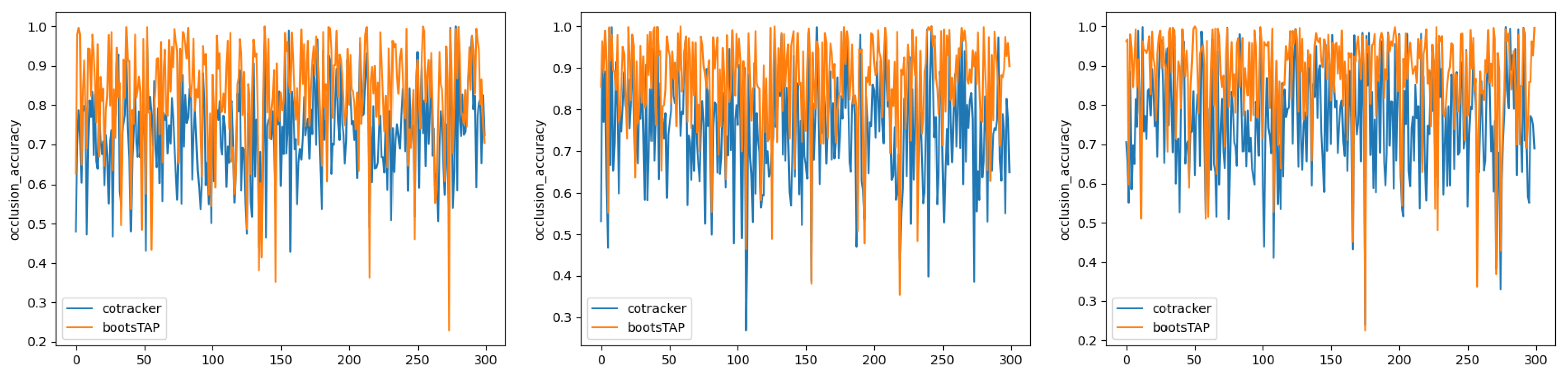}
    \caption{Comparison of occlusion accuracy}
    \label{fig:enter-label}
\end{figure}

\begin{figure}
    \centering
    \includegraphics[scale=0.3]{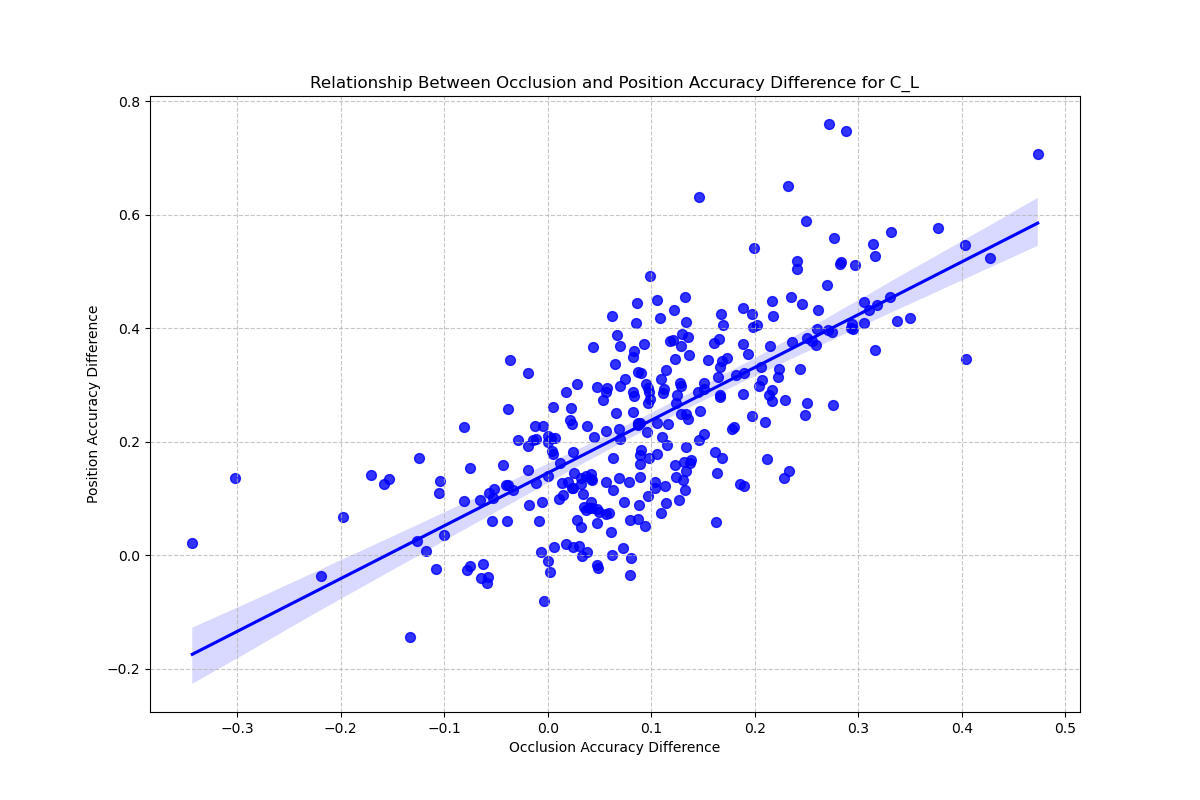}
    \caption{Correlation between occlusion accuracy and point position accuracy}
    \label{fig:enter-label}
\end{figure}

\end{appendix}

\clearpage
\bibliographystyle{unsrt}